\pdfoutput=1

\documentclass[11pt]{article}
\usepackage{amsmath}

\usepackage[final]{acl}

\usepackage{times}
\usepackage{latexsym}
\usepackage{xcolor}
\usepackage{multirow, array, booktabs, graphicx}
\usepackage{xspace}
\usepackage{tcolorbox}
\usepackage{tabularx}
\usepackage{booktabs}
\usepackage{longtable}
\usepackage{tabularx}
\usepackage{booktabs}
\usepackage{tabularx}
\usepackage{ragged2e}
\usepackage{makecell} 
\usepackage{array}
\usepackage{enumitem}
\usepackage{ragged2e}
\usepackage[normalem]{ulem}
\usepackage[table]{xcolor}
\tcbuselibrary{breakable}
\newcolumntype{C}{>{\centering\arraybackslash}X}

\usepackage[T1]{fontenc}

\usepackage[utf8]{inputenc}

\usepackage{microtype}

\usepackage{inconsolata}

\usepackage{graphicx}

\title{Text Analytics Evaluation Framework: \\ A Case Study on LLMs and Social Media}

\author{Yuefeng Shi \qquad
  Nedjma Ousidhoum  \qquad
  Jose Camacho-Collados \\
  School of Computer Science and Informatics, Cardiff University \\
  \texttt{\{ShiY49, OusidhoumN, CamachoColladosJ\}@cardiff.ac.uk}}

\begin{document}
\maketitle
\begin{abstract}

LLMs have demonstrated exceptional proficiency in a wide range of NLP tasks. However, a notable gap remains in practical data analysis scenarios, particularly when LLMs are required to process long sequences of unstructured documents, such as news feeds or, as specifically addressed in this paper, social media posts. To empirically assess the effectiveness of LLMs in this setting, we introduce a question-based evaluation framework comprising 470 manually curated questions designed to evaluate LLMs' semantic understanding and reasoning abilities over aggregated text data.
We apply our benchmark on diverse Twitter datasets covering various NLP tasks, including sentiment analysis, hate speech detection, and emotion recognition. Our results reveal that the performance depends heavily on input scale and the complexity of the data sources, declining noticeably in multi-label or target-dependent scenarios. In addition, as task complexity increases, performance drops progressively from basic semantic existence identification to more demanding operations such as comparison, counting, and calculation. Furthermore, as the input size grows beyond 500 instances, we identify a common limitation across LLMs, particularly Open-weights models: performance degrades substantially, especially on numerical tasks. These findings highlight critical architectural bottlenecks in current LLMs for performing rigorous quantitative analysis over large text collections.

\end{abstract}

\section{Introduction}

The rapid advancement of Large Language Models (LLMs)~\cite{achiam2023gpt, touvron2023llama, team2023gemini} has significantly expanded the scope of NLP applications, granting them the capacity to address a diverse array of complex, real-world tasks such as data analysis. However, unlike structured data, vast repositories of unstructured texts, ranging from customer service logs and financial earnings reports to clinical trial records, contain rich latent information that poses unique challenges, requiring not only complex semantic interpretation but also rigorous reasoning processes to aggregate the information effectively.

\begin{figure*}[t!]
    \centering
    \includegraphics[width=0.98\linewidth, keepaspectratio]{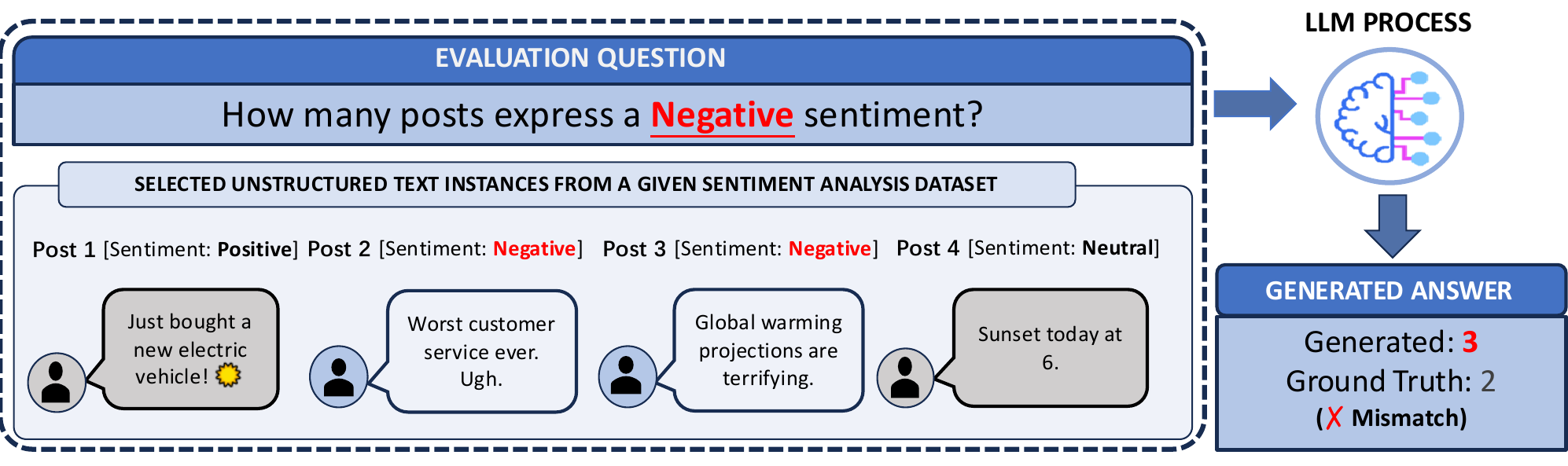}
    \caption{An illustrative example of our data analysis evaluation framework.}
    \label{fig:pipe}
\end{figure*}

Using LLMs as efficient, automated tools for data analysis can help us reduce the cost of deriving insights from massive texts. 
However, as we detail in the related work section, research targeting this setting remains
scarce, and lacks dedicated datasets and standardized
evaluation norms. Furthermore, previous research has confirmed that LLM
reliability is often compromised by hallucinations~\cite{schwartz2024numerologic}
and inadequate numerical and logical reasoning
capabilities~\cite{liu2025logical}. In data analysis tasks where
precision is a cornerstone requirement, whether LLMs continue to
exhibit these critical failures---and to what extent such limitations
hinder their ability to provide comprehensive and trustworthy
insights---remains an open question.

To address these gaps, we present a QA-based dataset designed for the preliminary exploration and evaluation of LLM abilities in analyzing unstructured text (See Figure~\ref{fig:pipe}). In particular, we focus on the data type encountered in social media, which is a domain that encapsulates our aims, given the short nature of social media text instances, and the need for analyzing large amounts of posts for different applications. Constructed on top of two well-established social media NLP benchmarks, TweetEval~\cite{barbieri2020tweeteval} and SuperTweetEval~\cite{antypas2023supertweeteval}, our dataset includes 470 manually filtered questions with human-verified ground truths. Our evaluation framework investigates three critical dimensions of model performance to provide a comprehensive assessment. First, we design questions of varying difficulty levels---ranging from simple true/false judgments to complex data computations---to test the depth of the models' analytical reasoning. Then, by incorporating text sources originally designed for diverse classification tasks, we evaluate the models' adaptability to different semantic complexities. Moreover, we use varying input data scales to examine model robustness when processing different volumes of textual information. Finally, we quantitatively benchmark various Open-weights and Closed-source LLMs to test their capabilities in real-world applications.
 
Crucially, we observe a clear capability gap heavily influenced by the semantic or structural complexity of the data source. In addition, while models are highly proficient in identifying the existence of specific semantics,
 their accuracy drops during comparisons, and they struggle with complex numerical ones such as counting and calculation. Additionally, we identified a critical turning point at approximately 500 input instances; beyond this threshold, Open-weights LLMs exhibit severe performance degradation across all tasks, and Closed-source models also struggle with numerical reasoning at this scale.

\section{Related Work}


Several benchmarks have been constructed to evaluate the data analysis capabilities of LLMs or LLM-based agents in relevant tasks~\cite{lai2022ds, he2024text2analysis}. However, the majority target structured data analysis. Prominent datasets such as InfiAgent-DABench~\cite{hu2024infiagent} and StatLLM~\cite{song2025statllm} use CSV files or statistical languages (e.g., SAS) to assess tasks like planning and code generation. Similarly, DataSciBench~\cite{zhang2025datascibench} and DABstep~\cite{egg2025dabstep} focus on executing complex data science workflows via executable code. While these benchmarks effectively test quantitative reasoning on structured tables, they do not address the challenges of analyzing unstructured text and do not evaluate tasks that demand deep semantic understanding. Another direction for data analysis using LLMs takes a system-centric perspective. UDA-Bench~\cite{deng2025unstructured} focusing on query interfaces and operator design for SQL-like querying over unstructured data, rather than evaluating LLM reasoning capabilities directly. 

In contrast, traditional NLP benchmarks such as document-level Question Answering~\cite{rasool2024evaluating} and long-context aggregation benchmarks~\cite{shaham2022scrolls} target factual
retrieval or synthesis within a single document, and do not address the challenge of analyzing a collection of short, noisy, independent instances. Instead, our task requires models to first perform implicit semantic inference across all instances, and then aggregate the result through reasoning. 

Most closely related to our work are Oolong~\cite{bertsch2025oolong}, which evaluates
cross-instance aggregation over long contexts, and \citet{chen2026mip}, who study LLM performance degradation under multi-instance processing.
However, both use data with relatively straightforward semantics, whereas our benchmark is built on social media text carrying more complicated semantic information---such as mixed emotions or implicit stances---that models can easily misinterpret at the instance level before any aggregation begins. Furthermore, beyond simple counting or
percentage computation, our questions involve first-order logical reasoning and cross-category comparison, posing a greater and more diverse inferential challenge than prior work.

\section{Text Analytics Evaluation Framework}

 We aim to construct an evaluation pipeline and a benchmark to assess the data analysis capabilities of LLMs on unstructured textual data\footnote{We release the evaluation framework at \url{https://github.com/yuefeng-shi/text_data_analysis}
}. The benchmark focuses on both real-world textual application scenarios in social media and the intrinsic statistical properties of the data. Among others, this framework can help evaluate the following abilities of LLMs:
interpreting data distributions,
understanding simple logic,
and performing mathematical computations relevant to data analysis.

\subsection{Task Definition}
We define the task as a generative document-based question answering problem~\cite{fan2019eli5}. Formally, let $\mathcal{D} = \{ d_i \}_{i=1}^N$ denote a text dataset, where $d_i$ represents an individual text instance. Given an input question $q$ related to data analysis, the question answering process is formalized as:

\begin{equation}
    a = \mathcal{M}_{\text{LLM}} \left( q, \mathcal{D} \right)
    \label{eq:gen_qa}
\end{equation}

\noindent where $a$ denotes the generated answer, and $\mathcal{M}_{\text{LLM}}$ is the LLM-based system that internally derives statistical information from $\mathcal{D}$ to generate $a$.

\subsection{Dataset Construction}

\paragraph{Data Selection}

We use social media posts as the primary data source due to their accessibility and widespread use in data analytics across domains, such as business intelligence and public opinion monitoring. 
We build our dataset using the publicly available benchmarks TweetEval~\cite{barbieri2020tweeteval} \textbf{(TE)} and its extension, SuperTweetEval~\cite{antypas2023supertweeteval} \textbf{(STE)}, which are available in English only. These benchmarks aggregate several high-quality datasets including diverse tasks and domains, such as sentiment analysis, emotion recognition 
or topic classification. 
We integrate eight datasets from TE and STE, which we systematically categorize into the following groups based on annotation complexity (see Table~\ref{tab:datasets_grouped}):

\begin{enumerate}[leftmargin=*]
\item \textbf{Group 1: Single-Label Classification (SLC)} includes sentiment analysis \textbf{(TE-SA)}, hate speech detection \textbf{(TE-HS)}, offensive language detection \textbf{(TE-OL)} and hate speech detection (\textbf{STE-HS}). These tasks share a small and mutually exclusive label space. This represents a lower structural complexity at the output level compared to tasks with richer or overlapping categorical signals.
\item \textbf{Group 2: Multi-Label Classification (MLC)} includes emotion recognition (\textbf{STE-ER}) and topic classification (\textbf{STE-TC}). These tasks are more complex, as they require handling overlapping and non-mutually exclusive categories.
\item \textbf{Group 3: Targeted Classification (TC)} includes stance detection (\textbf{TE-SD}) and targeted sentiment analysis (\textbf{STE-SA}). Unlike standard classification tasks, these involve an explicit target variable and require deeper contextual reasoning to determine the relationship between the text and a specific entity. These tasks allow us to assess the model's contextual adaptability.
\end{enumerate}

\begin{table}[t]
\centering
\renewcommand{\arraystretch}{1.2}
\resizebox{\columnwidth}{!}{
\begin{tabular}{ll}
\toprule
\textbf{Type} & \textbf{Dataset Name} \\ 
\midrule
\multicolumn{2}{l}{\textit{\textbf{Single-Label Classification (SLC)}}} \\
 & TE-Sentiment Analysis (TE-SA)~\cite{rosenthal2017semeval} \\
 & TE-Hate Speech Detection (TE-HS)~\cite{basile-etal-2019-semeval} \\
 & TE-Offensive Language Detection (TE-OL)~\cite{zampieri2019semeval} \\
 & STE-Hate Speech Detection (STE-HS)~\cite{sachdeva-etal-2022-measuring} \\
\midrule
\multicolumn{2}{l}{\textit{\textbf{Multi-Label Classification (MLC)}}} \\
 & STE-Emotion Recognition (STE-ER)~\cite{mohammad-etal-2018-semeval} \\
 & STE-Topic Classification (STE-TC)~\cite{antypas-etal-2022-twitter} \\
\midrule
\multicolumn{2}{l}{\textit{\textbf{Targeted Classification}}} \\
 & TE-Stance Detection (TE-SD)~\cite{mohammad2016semeval} \\
 & STE-Sentiment Analysis (STE-SA)~\cite{rosenthal-etal-2017-semeval} \\
\bottomrule
\end{tabular}
}
\caption{Categorization of Datasets by Task Type. TE denotes TweetEval, and STE denotes SuperTweetEval.}
\label{tab:datasets_grouped}
\vspace*{-3mm}
\end{table}

\paragraph{Question Categorization}

To align our evaluation framework with real-world data analysis workflows and assess specific limitations in current LLM capabilities, we design evaluation questions covering four main types, reflecting different levels of reasoning demand:

\begin{enumerate}[leftmargin=*, noitemsep]
    \item \textbf{Existence} questions test the model's precision in detecting the presence or absence of specific labels, explicitly assessing hallucination. (e.g. 'Does the dataset contain entries expressing the 'anger' emotion?' in STE-ER)   
    \item \textbf{Count} questions evaluate the model's ability to maintain global attention and perform accurate counting within long textual contexts (e.g. 'How many entries do not express the 'trust' emotion?' in STE-ER)  
    \item \textbf{Comparison} questions assess logical reasoning by requiring the model to retain multiple frequency counts in working memory and perform binary comparisons between label groups. (e.g. 'Do entries expressing 'love' emotion account more than 30\% of the total dataset?' in STE-ER)
    \item \textbf{Calculation} questions require deriving secondary metrics (e.g., percentages) from raw text, thereby testing model's mathematical reasoning. (e.g. 'Calculate the percentage of entries with both 'sadness' and 'anger' emotions.' in STE-ER)
\end{enumerate}

\begin{table*}[htbp]
    \centering
    \small 
    \renewcommand{\arraystretch}{1.2} 

    \begin{tabularx}{\textwidth}{@{} l l X c @{}}
        \toprule
        \textbf{Answer Category} & \textbf{Question Category} & \textbf{Question} & \textbf{Answer} \\
        \midrule
        \multirow{2}{*}{\textbf{Binary (Yes/No)}} & Existence & Does the dataset contain negative sentiment entries? & Yes \\
         & Comparison & Is the proportion of negative sentiment below 30\%? & No \\
        \midrule
        \multirow{2}{*}{\textbf{Numerical}} & Count & What is the number of positive sentiment instances? & 24 \\
         & Calculation & What is the combined count of hateful and not hateful entries? & 100 \\
        \midrule
        \textbf{Proportional} & Calculation & What is the proportion of instances that are neutral? & 35\% \\
        \bottomrule
    \end{tabularx}
    \caption{Examples of generated questions categorized by answer format and question type. }
    \label{tab:question_examples}
    \vspace*{-3mm}
\end{table*}

Based on these tasks, we categorize the required outputs into three distinct groups based on the required \textbf{answer format}:

\begin{enumerate}[leftmargin=*]

\item  \textbf{Binary Response (Yes/No):} This category requires the model to make logical judgments based on textual evidence. It corresponds to \textit{Existence} and \textit{Comparison} questions.

\item  \textbf{Numerical Response (Number):} This category focuses on both frequency quantification and computational complexity. It requires the model not only to accurately count specific instances within the dataset, but also to process fundamental math operations, directly mapping to \textit{Count} and \textit{Calculation} questions.
    
\item  \textbf{Proportional Response (Percentage):} This category requires the model to compute relative metrics, necessitating explicit arithmetic operations , typically found in \textit{Calculation} questions.

\end{enumerate}

\subsection{Evaluation Pipeline}

\paragraph{Data Sampling and Preprocessing} 
To evaluate model capabilities across varying information volumes, we applied a random sampling strategy to the source datasets, generating subsets at seven scales: 10, 50, 100, 250, 500, 750, and 1,000 instances. The sampling was carried out to preserve the original label distribution to ensure similar statistical representativeness from all labels, and three independent samples were drawn for each scale to mitigate random variance. To simulate a purely unstructured text analysis environment, we only kept the raw text and removed labels and meta-data.

\paragraph{Question Generation and Filtering}

We first used GPT-4~\cite{achiam2023gpt} to generate candidate questions for each task, guided by prompts that enforced relevance to real-world data analysis scenarios and required all questions to be grounded in the task-specific label space, producing 10 Binary (Yes/No) and 10 Numerical (Number/Percentage) questions per task (see Appendix~\ref{sec:Datagenprompts}). 
We then manually removed unsuitable candidates, corrected grammatical issues, and consolidated the remaining ones into shared sentence templates by merging and complementing questions across tasks, so as to ensure both coverage of all four question categories---Existence, Count, Comparison, and Calculation---and sufficient diversity in question formulation. Additionally, we further designed questions by hand to cover label co-occurrence statistics and cross-target comparisons for multi-label classification and target-based tasks.
Finally, we instantiated the full question set from these templates. Since template-based generation tends to produce formulaic phrasing, each question was manually post-edited to improve naturalness. The modification rate was 68\% across all questions, although most modifications were rather minor.

\paragraph{Ground Truth Verification}
All ground truth answers were deterministically computed using custom Python scripts applied to the sampled data. Each answer was independently verified by the authors to ensure correctness. To prevent clustering around specific values, we adjusted the distribution of numerical answers during filtering, and rebalanced questions by explicitly shifting the target label focus where necessary. Table~\ref{tab:question_examples} presents examples of 
questions alongside their verified answers.

\paragraph{Benchmark Statistics}

Our final benchmark covers eight datasets, each sampled into seven size configurations ranging from 10 to 1,000 instances. As detailed in \autoref{tab:merged_statistics}, the benchmark comprises 470 rigorously validated question-answer pairs with an average length of 11.6 words. The table further breaks down the distribution along two major dimensions: 1)\ Answer Format and 2)\ Logical Category. Binary (Yes/No) responses constitute 61.5\% of the dataset (n=289), serving as the primary evaluation format, while Numerical ($n=91$) and Proportional 
($n=90$) responses each account for approximately 19\% of the benchmark. In terms of logical complexity, the foundational tasks---Existence ($n=83$) and Count ($n=61$)---provide a baseline, whereas higher-order 
reasoning tasks---Calculation ($n=120$) and Comparison ($n=206$)---form the majority at approximately 69\%, reflecting the benchmark's focus on deep analytical capabilities. Detailed statistics are provided in 
Appendix~\ref{tab:fulldatastas}.

\begin{table}[t]
\centering
\footnotesize 
\setlength{\tabcolsep}{2pt} 
\renewcommand{\arraystretch}{1.15} 

\begin{tabularx}{\columnwidth}{X r} 
\toprule
\textbf{Metric / Category} & \textbf{Value / Count} \\
\midrule

\multicolumn{2}{l}{\textit{\textbf{Dataset Overview}}} \\
Total Datasets & 8\\
\hspace{1em} Single-label Classification & 4 \\
\hspace{1em} Multi-label Classification & 2 \\
\hspace{1em} Targeted Classification & 2 \\
\addlinespace[3pt] 
Dataset Sample Sizes & \parbox[t]{3cm}{\raggedleft 10, 50, 100, 250, \\ 500, 750, 1000} \\

\midrule

\multicolumn{2}{l}{\textit{\textbf{Question Statistics (Total: 470)}}} \\
Avg. Question Length & 11.6 words \\

\addlinespace[2pt] 
\textit{By Answer Format:} & \\
\hspace{1em} Binary (Yes/No) & 289 \\
\hspace{1em} Numerical (Number) & 91 \\
\hspace{1em} Proportional (Percentage) & 90 \\

\addlinespace[2pt] 
\textit{By Logical Category:} & \\
\hspace{1em} Calculation & 120 \\
\hspace{1em} Comparison & 206 \\
\hspace{1em} Count & 61 \\
\hspace{1em} Existence & 83 \\

\bottomrule
\end{tabularx}
\caption{Consolidated statistics of the benchmark datasets and questions.}
\label{tab:merged_statistics}
\end{table}

\section{Experimental Setting}

In this section, we detail our main experimental settings---the models analyzed,  prompting strategies, and evaluation metrics to evaluate their adaptability and instruction-following capabilities.

\subsection{Comparison Models}

\noindent \textbf{Closed-source Models.} This group consists of commercial models often achieving state-of-the-art in NLP tasks and beyond. We use Gemini-3.1-Flash-Lite~\cite{gemini31lite}, GPT-5.4-Mini~\cite{gpt54mini}, and Grok-4.1-Fast~\cite{grok41fast}.\footnote{We apply cost-effective solutions rather than large reasoning models (e.g., GPT-5.4) because our evaluation is conducted under a non-reasoning setting and a fixed budget.} 

\noindent \textbf{Open-weights Models.} To assess the capabilities of accessible weights across different parameter scales, we test different model families: LLaMA and Qwen. From the LLaMA series~\cite{grattafiori2024llama}, we select models from the lightweight LLaMA-3.2-3B-Instruct, the mid-sized LLaMA-3.1-8B-Instruct, and the large-sized LLaMA-3.3-70B-Instruct. Similarly, for Qwen~\cite{yang2025qwen3} dense models, we include Qwen-3.5-27B-Instruct, Qwen-3.5-9B-Instruct, and Qwen-3-4B-Instruct. This selection enables a systematic analysis of how performance scales with model size. We also apply Qwen Mixture-of-Experts (MoE) models~\cite{qwen35} Qwen-3.5-397B-A17B and Qwen-3.5-122B-A10B.

\subsection{Prompting} 
\label{sec:mainpaperprompt}
To evaluate the models' intrinsic analytical capabilities, we provide each model with a prompt containing the task description, data source metadata, and complete label taxonomy,  while enforcing strict output constraints (Yes/No, Number, or Percentage without any explanations). Full prompt templates are provided in Appendix~\ref{sec:taskprompt}.

\subsection{Evaluation Metrics}
\label{sec:evaluation_metrics}


We evaluate model performance across three distinct task categories, using metrics specifically tailored to the output nature of each question type:

\paragraph{Classification Tasks.}
For yes/no output questions, we report the \textbf{Macro-F1 score}. F1 provides a balanced assessment of precision and recall, ensuring robustness against potential class imbalances.

\paragraph{Percentage Regression.}
For percentage output questions, we implement the Range-Normalized Root Mean Squared Error (RNRMSE). Although the theoretical range of percentage values is $[0, 1]$, the empirical distribution of gold labels is relatively narrower in practice, meaning that standard MSE would systematically understate the relative severity of prediction errors. Therefore, we normalize each error by the observed
gold label range $R$, so that errors are measured relative to the actual spread of values the model is expected to discriminate:
\begin{align}
    \text{RNRMSE} &= \sqrt{\frac{1}{n} \sum_{i=1}^{n}
        \left( \frac{y_i - \hat{y}_i}{R} \right)^2} \nonumber\\
    \text{where } R &= \max(y) - \min(y)
\end{align}
Here, $R$ serves as the normalization factor calibrated to the true label
distribution, allowing for an interpretable assessment of prediction
quality regardless of how concentrated the gold values are.

\paragraph{Numerical Regression.}
For numerical output questions, we implement the Normalized Root-Mean-Square Error (NRMSE). Despite being a regression task, strictly absolute errors in numerical prediction can be misleading due to the wide variance in ground truth magnitudes (e.g., from tens to thousands). Therefore, we scale the penalty relative to the specific data size ($S_i$) of each sample. This ensures that errors in large-scale datasets do not disproportionately dominate the metric:
\begin{equation}
    \text{NRMSE} = \sqrt{ \frac{1}{N} \sum_{i=1}^{N} \left( \frac{y_i - \hat{y}_i}{S_i} \right)^2 }
\end{equation}
Here, $S_i$ serves as the sample-specific normalization factor (e.g., the total dataset size corresponding to the query), allowing for a fair comparison across varying instance counts as we explore in our evaluation settings.

\begin{table*}[!t]
\centering
\renewcommand{\arraystretch}{0.85}   
\setlength{\tabcolsep}{2.0pt}        
\footnotesize
\begin{tabular}{@{}c l *{4}{c} | *{2}{c} | *{2}{c}@{}} 
\toprule
\textbf{Metric} & \textbf{Model} & TE-SA & TE-HS & TE-OL & STE-HS & STE-ER & STE-TC & STE-SA & TE-SD \\
\midrule
\multirow{12}{*}{F1 $\uparrow$} 
  & \textit{Baseline} & 0.335 & 0.333 & 0.333 & 0.347 & 0.343 & 0.338 & 0.341 & 0.342 \\
\cmidrule{2-10}
  & \multicolumn{9}{l}{\small\textit{Closed-source}} \\
  & GPT-5.4-Mini & 0.687 & 0.621 & 0.526 & 0.577 & 0.726 & 0.696 & 0.630 & 0.638 \\
  & Gemini-3.1-Flash-Lite & 0.703 & \textbf{0.687} & 0.646 & 0.736 & \textbf{0.770} & 0.782 & 0.651 & 0.622 \\
  & Grok-4.1-Fast & 0.773 & 0.601 & 0.569 & 0.720 & 0.658 & \textbf{0.834} & 0.648 & 0.666 \\
\cmidrule{2-10}
  & \multicolumn{9}{l}{\small\textit{Open-weights}} \\
  & Qwen-3.5-397B-A17B & \textbf{0.795} & 0.676 & 0.617 & 0.772 & 0.677 & 0.795 & 0.678 & \textbf{0.704} \\
  & Qwen-3.5-122B-A10B & 0.722 & 0.579 & 0.566 & \textbf{0.791} & 0.703 & 0.767 & \textbf{0.747} & 0.659 \\
  & Qwen-3.5-27B-Instruct & 0.779 & 0.676 & \textbf{0.682} & 0.745 & 0.653 & 0.724 & 0.708 & 0.636 \\
  & LLaMA-3.3-70B-Instruct & 0.665 & 0.519 & 0.515 & 0.560 & 0.699 & 0.721 & 0.595 & 0.640 \\
  & LLaMA-3.1-8B-Instruct & 0.655 & 0.604 & 0.551 & 0.570 & 0.616 & 0.623 & 0.618 & 0.613 \\
  & Qwen-3.5-9B-Instruct & 0.681 & 0.521 & 0.431 & 0.708 & 0.668 & 0.739 & 0.678 & 0.689 \\
  & Qwen-3-4B-Instruct & 0.337 & 0.372 & 0.284 & 0.230 & 0.360 & 0.281 & 0.277 & 0.344 \\
  & LLaMA-3.2-3B-Instruct & 0.385 & 0.520 & 0.463 & 0.446 & 0.365 & 0.431 & 0.521 & 0.442 \\
\midrule
\multirow{12}{*}{RNRMSE $\downarrow$}
  & \textit{Baseline} & 0.346 & 0.287 & 0.335 & 0.388 & 0.356 & 0.401 & 0.393 & 0.374 \\
\cmidrule{2-10}
  & \multicolumn{9}{l}{\small\textit{Closed-source}} \\
  & GPT-5.4-Mini & 0.294 & 0.345 & 0.423 & 0.315 & \textbf{0.260} & 0.272 & 0.445 & 0.517 \\
  & Gemini-3.1-Flash-Lite & \textbf{0.178} & \textbf{0.142} & \textbf{0.215} & \textbf{0.252} & 0.332 & 0.262 & \textbf{0.343} & 0.429 \\
  & Grok-4.1-Fast & 0.304 & 0.369 & 0.409 & 0.268 & 0.346 & 0.299 & 0.408 & 0.486 \\
\cmidrule{2-10}
  & \multicolumn{9}{l}{\small\textit{Open-weights}} \\
  & Qwen-3.5-397B-A17B & 0.251 & 0.232 & 0.323 & 0.257 & 0.348 & \textbf{0.223} & 0.361 & \textbf{0.284} \\
  & Qwen-3.5-122B-A10B & 0.409 & 0.330 & 0.368 & 0.323 & 0.293 & 0.314 & 0.405 & 0.429 \\
  & Qwen-3.5-27B-Instruct & 0.312 & 0.186 & 0.255 & 0.330 & 0.375 & 0.284 & 0.353 & 0.399 \\
  & LLaMA-3.3-70B-Instruct & 0.826 & 0.628 & 0.791 & 0.754 & 0.725 & 0.719 & 0.669 & 0.730 \\
  & LLaMA-3.1-8B-Instruct & 0.839 & 0.753 & 0.809 & 0.805 & 0.663 & 0.770 & 0.724 & 0.736 \\
  & Qwen-3.5-9B-Instruct & 0.284 & 0.505 & 0.560 & 0.474 & 0.358 & 0.323 & 0.381 & 0.435 \\
  & Qwen-3-4B-Instruct & 0.720 & 0.661 & 0.825 & 0.749 & 0.596 & 0.769 & 0.729 & 0.684 \\
  & LLaMA-3.2-3B-Instruct & 0.493 & 0.743 & 0.826 & 0.886 & 0.507 & 0.591 & 0.462 & 0.413 \\
\midrule
\multirow{12}{*}{NRMSE $\downarrow$}
  & \textit{Baseline} & 0.257 & 0.290 & 0.308 & 0.352 & 0.359 & 0.330 & 0.361 & 0.340 \\
\cmidrule{2-10}
  & \multicolumn{9}{l}{\small\textit{Closed-source}} \\
  & GPT-5.4-Mini & \textbf{0.180} & 0.307 & 0.320 & 0.298 & 0.299 & \textbf{0.185} & 0.252 & \textbf{0.222} \\
  & Gemini-3.1-Flash-Lite & 0.194 & \textbf{0.237} & \textbf{0.268} & \textbf{0.277} & \textbf{0.297} & 0.228 & 0.255 & 0.227 \\
  & Grok-4.1-Fast & 0.206 & 0.343 & 0.336 & \textbf{0.277} & 0.323 & 0.197 & \textbf{0.242} & \textbf{0.222} \\
\cmidrule{2-10}
  & \multicolumn{9}{l}{\small\textit{Open-weights}} \\
  & Qwen-3.5-397B-A17B & 0.260 & 0.298 & 0.334 & 0.301 & 0.334 & 0.234 & 0.270 & 0.251 \\
  & Qwen-3.5-122B-A10B & 0.266 & 0.301 & 0.335 & 0.322 & 0.333 & 0.239 & 0.256 & 0.245 \\
  & Qwen-3.5-27B-Instruct & 0.240 & 0.331 & 0.382 & 0.290 & 0.332 & 0.239 & 0.271 & 0.242 \\
  & LLaMA-3.3-70B-Instruct & 0.576 & 0.566 & 0.576 & 0.588 & 0.564 & 0.483 & 0.393 & 0.451 \\
  & LLaMA-3.1-8B-Instruct & 0.585 & 0.702 & 0.596 & 0.568 & 0.464 & 0.462 & 0.447 & 0.462 \\
  & Qwen-3.5-9B-Instruct & 0.318 & 0.406 & 0.440 & 0.385 & 0.360 & 0.290 & 0.262 & 0.272 \\
  & Qwen-3-4B-Instruct & 0.758 & 0.780 & 0.781 & 0.758 & 0.706 & 0.758 & 0.670 & 0.609 \\
  & LLaMA-3.2-3B-Instruct & 0.595 & 0.673 & 0.630 & 0.708 & 0.528 & 0.557 & 0.463 & 0.646 \\
\bottomrule
\end{tabular}

\caption{Main results across the eight TE/STE benchmarks. \textbf{F1}: Macro-F1 ($\uparrow$ higher is better), RNRMSE and NRMSE ($\downarrow$ lower is better). \textbf{Bold} indicates the best non-baseline score per dataset. \textit{Baseline} denotes the best constant predictor selected from multiple heuristic strategies (e.g. all questions set yes, 50\% or 10 as answers).  }
\label{tab:main_results}
\end{table*}

\section{Results}

The evaluation results are summarized in Table~\ref{tab:main_results}. Each entry reports the mean performance aggregated across all data sizes.

\subsection{Binary Classification Questions}

\noindent \textbf{Model performance.} Closed-source models generally outperform Open-weights models on binary classification across the board, with Grok-4.1-Fast reaching 0.834 on STE-TC. Nevertheless, larger Qwen-3.5 models perform competitively with Closed-source models. 
For example, Qwen-3.5-397B-A17B achieves 0.795 on TE-SA, outperforming the best Closed-source model on this task (Grok-4.1-Fast at 0.773). Similarly, Qwen-3.5-122B-A10B reaches 0.791 on STE-HS, surpassing Gemini-3.1-Flash-Lite (0.736).
Most models above 9B exceed the constant-prediction baseline (0.333–0.347), but this does not hold for smaller models. Qwen-3-4B-Instruct performs below random scores in some datasets, only reaching 0.230 on STE-HS against a baseline of 0.347, and LLaMA-3.2-3B-Instruct follows a similar pattern on several datasets. STE-HS is also worth noting here: its questions carry higher semantic complexity than other SLC datasets, involving multi-step reasoning over fine-grained hate sub-categories, and this appears to challenge all model tiers beyond what the simple classification framing might suggest.

\noindent \textbf{Data complexity.} Targeted data source (STE-SA and TE-SD) on Binary Classification tasks drops dramatically compared to SLC and MLC in performance because their dual-input structure forces the model to conditionally bind the target to the context rather than simply classifying the text. This added semantic difficulty explains why top models like Qwen-3.5-397B-A17B drop significantly from 0.795 on TE-SA to 0.704 on TE-SD. Within SLC data source itself, dataset-level difficulty also varies considerably: the gap between TE-SA and STE-HS highlights that fine-grained semantic distributions can challenge models just as much as structural complexity.

\begin{figure*}[t!]
    \centering
    \includegraphics[width=0.95\linewidth, keepaspectratio]{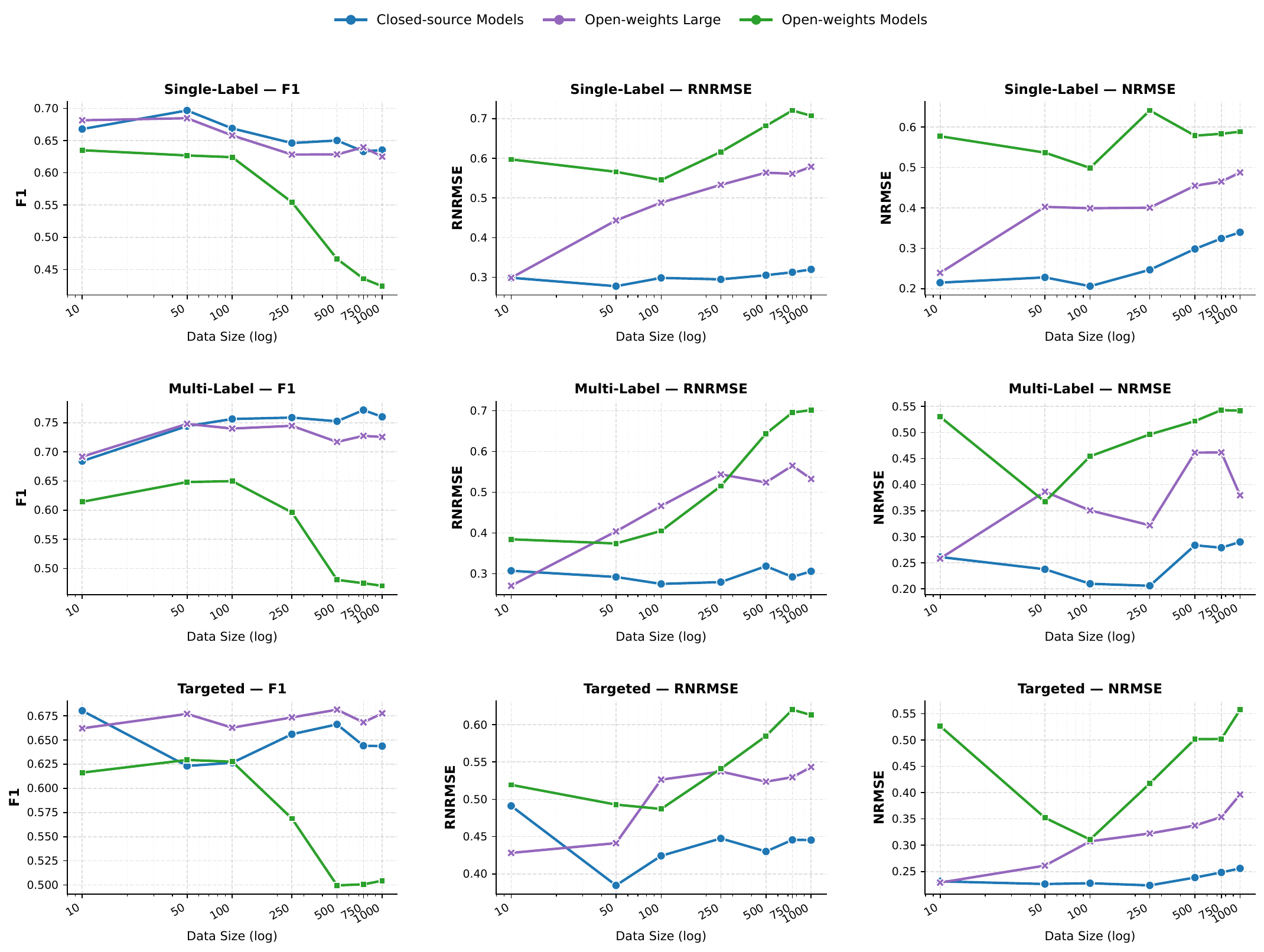}
    \caption{Performance across data sizes by metric and data category. Results are averaged across types of model. 'Open-weights Large' includes Qwen-3.5-397B-A17B, Qwen-3.5-122B-A10B and  LLaMA-3.3-70B-Instruct, while the remaining Open-weights models are included in the general category. 
    }
    \vspace*{-2mm}
    \label{fig:datasizefmse}
\end{figure*}
\subsection{Percentage Estimation Questions} 
\noindent \textbf{Model performance.} For the percentage task, Closed-source models maintain a clear advantage in percentage estimation, with Gemini-3.1-Flash-Lite setting the benchmark by achieving the lowest RNRMSE on five of eight datasets. In contrast, a significant portion of the evaluated models fail to even match the random prediction baseline. This barrier is most evident in models under 9B, especially for LLaMA-3.2-3B-Instruct and Qwen-3-4B-Instruct, which yield estimates that often under perform simple heuristics. Even more, larger LLaMA family models like LLaMA-3.3-70B-Instruct perform below the baseline performance (0.628–0.826), indicating that the entire LLaMA series struggle with this specific type of percentage task.

\noindent \textbf{Data complexity.} Percentage estimation becomes harder as task complexity increases. For most models, the lowest prediction error is on SLC data source, moderate on MLC data source, and highest on Targeted data source. Targeted data source produces the largest errors; for example, GPT-5.4-Mini scores 0.294–0.423 on SLC data sources but its error rises to 0.445–0.517 on Targeted ones. This is because Targeted questions ask for percentages within highly specific groups with the complex dual-input nature (a specific entity to a specific semantic) at the same time. Since both the reference group and the target category are harder to identify, the prediction errors naturally increase. The error rates for MLC data source fall between SLC and Targeted ones, which matches the complexity trend observed in classification.

\subsection{Numerical Regression Questions}
\noindent \textbf{Model performance.} Closed-source models again lead on numerical regression tasks, and the gap to Open-weights models is more pronounced here than on classification. GPT-5.4-Mini and Gemini-3.1-Flash-Lite perform the best, with GPT reaching 0.180 on TE-SA and 0.185 on STE-TC, and Gemini hitting 0.237 on TE-HS and 0.268 on TE-OL. Large Qwen-3.5 variants hold up reasonably well, staying within 0.240–0.382 on most SLC and MLC datasets, but LLaMA-3.2-3B (0.463–0.708) and Qwen-3-4B (0.609–0.781) fail to beat the baseline in most cases. The overall picture is clear: numerical reasoning capabilities break down sharply in smaller Open-weights models, making them ineffective at executing such tasks.

\noindent \textbf{Data complexity.} Errors generally rise from SLC to MLC, which is consistent with the pattern seen across tasks. However, unlike other tasks, Targeted data sources show lower NRMSE than SLC across all model tiers. 
This is mostly because of how those datasets are constructed: Targeted questions ask about a specific target entity, so gold labels to questions naturally represent a small fraction of the total dataset size. Since models tend to underestimate counts, those smaller gold values produce lower normalized errors automatically, rather than the task difficulty. 
Moreover, it is also worth noting that format validity is noticeably lower on some SLC data source such as STE-HS, where number-format validity drops to 84.7\% and percentage-format validity to 77.3\%, compared to 91–94\% and 84–87\% respectively on MLC and Targeted datasets (Appendix~\ref{sec:Generrstas}). This translates to a larger NRMSE penalty for SLC than for MLC.

\section{Analysis}

In addition to the overall results presented in the previous section, we provide a detailed analysis by data size, prompt strategy and question type.

\subsection{Data Size Analysis}
We further analyze the performance trends of different model categories across varying input data sizes, conditioned on different data categories and model types. 
Results are shown in Figure~\ref{fig:datasizefmse}. Overall, there is a clear negative correlation between the number of input instances and model performance, with numerical aggregation performance decreasing once the input exceeds 500 samples across all model types.

While most models achieve peak performance on a smaller set of inputs (10–50 samples), their performance trends begin to contrast sharply as the sample size increases. In Binary Response tasks, Closed-source and large Open-weights models experience only minimal decline as the scale grows, whereas standard Open-weights models experience a sharp decline and ultimately collapse. However, despite the semantic robustness of advanced models, processing a massive volume of independent instances exposes a universal limitation in numerical reasoning. Once the input exceeds the 500-sample threshold, error rates (RNRMSE and NRMSE) increase across all models. Even the most capable models suffer from increasing numerical errors as the input approaches 1,000 samples, highlighting a shared architectural bottleneck among current LLMs in aggregating quantitative data over large collections of text.

\subsection{Prompt Strategy Analysis}

\begin{table}[htbp] 
\centering
\resizebox{\columnwidth}{!}{
\begin{tabular}{llccc}
\toprule
\textbf{Metric} & \textbf{Model} & \textbf{No Instr.} & \textbf{Standard} & \textbf{One-Shot} \\
\midrule
\multirow{2}{*}{F1 $\uparrow$}
  & Closed-source & 0.673 & 0.673 & \textbf{0.711} \\
  & Open-weights  & 0.641 & 0.690 & \textbf{0.740} \\
\midrule
\multirow{2}{*}{RNRMSE $\downarrow$}
  & Closed-source & 0.479 & \textbf{0.396} & 0.433 \\
  & Open-weights  & 0.530 & 0.471 & \textbf{0.398} \\
\midrule
\multirow{2}{*}{NRMSE $\downarrow$}
  & Closed-source & 0.338 & \textbf{0.324} & 0.358 \\
  & Open-weights  & 0.378 & 0.350 & \textbf{0.277} \\
\bottomrule
\end{tabular}%
}
\caption{Comparison of prompting strategies across task types and models. The best performing strategy is highlighted in \textbf{bold}. (No Instr. = No Instruction)}
\label{tab:prompt_strategy}
\end{table}

To further evaluate the impact of prompt strategies, we sampled 120 question instances and tested three prompt strategies: NoInstruction, Standard, and One-Shot. "Standard" is the prompt introduced in Section~\ref{sec:mainpaperprompt}. Based on this, we introduce a "NoInstruction" setting, which removes the data source descriptions from the Standard and a "One-Shot" setting, which adds a designed input-output example. LLMs aggregated results are shown in Table~\ref{tab:prompt_strategy}. 

Experiments reveal that Open-weights models benefit from more complex prompts. Across all tasks, performance consistently improves from NoInstruction to Standard, and further to One-Shot prompting. However, this positive trend is not general. In numerical and percentage tasks, the One-Shot prompt actually performs worse using Closed-source models, leading to a noticeable performance decrease over different tasks. Ultimately, when averaging the performance across different model types and task categories, the Standard prompt emerges as the most balanced and robust strategy overall, successfully avoiding the severe performance drops compared to the other two variations.

\subsection{Analysis by Question Type}

\begin{table}[t]
    \centering
    \setlength{\tabcolsep}{2.2pt} 
    
    \resizebox{\columnwidth}{!}{
    \begin{tabular}{l ccc ccc}
    \toprule
    \multirow{2}{*}{\textbf{Task (Metric)}} & \multicolumn{3}{c}{\textbf{Closed-source}} & \multicolumn{3}{c}{\textbf{Open-weights}} \\
    \cmidrule(lr){2-4} \cmidrule(lr){5-7}
     & \textbf{SLC} & \textbf{MLC} & \textbf{Targeted} & \textbf{SLC} & \textbf{MLC} & \textbf{Targeted} \\
    \midrule
    
    Existence (F1) $\uparrow$        & 0.887 & 0.873 & 0.799 & 0.787 & 0.753 & 0.726 \\
    Comparison (F1) $\uparrow$      & 0.553 & 0.680 & 0.552 & 0.492 & 0.553 & 0.530 \\
    Calculation (RNRMSE) $\downarrow$   & 0.234 & 0.279 & 0.419 & 0.449 & 0.463 & 0.491 \\
    Calculation (NRMSE) $\downarrow$ & 0.301 & 0.303 & 0.151 & 0.547 & 0.455 & 0.359 \\
    Count (NRMSE) $\downarrow$       & 0.255 & 0.249 & 0.283 & 0.457 & 0.424 & 0.400 \\
    
    \bottomrule
    \end{tabular}
    }
    
    \caption{Performance breakdown by model category across Single-label Classification (SLC), Multi-label Classification (MLC), and Targeted Classification (Targeted) data sources. All reported values represent the average performance aggregated across multiple models within each category and across various data sources and sizes.}
    \vspace*{-3mm}
    \label{tab:task_breakdown}
\end{table}

Table \ref{tab:task_breakdown} shows the performance across four question types under different data source complexities. These results reflect the average performance aggregated across all evaluated models within each category and across multiple dataset sources and sizes. This breakdown demonstrates that tasks become harder as they require deeper reasoning. Models perform well on simple Existence questions. The highest F1 score reaches an impressive 0.887 on SLC, and even the lowest score in this category (in the Targeted context) exceeds 70\% (0.726), which corroborates current LLMs' strong ability to identify the existence of specific semantics. However, in stark contrast to the strong results on Existence tasks, performance drops dramatically on Comparison questions. Closed-source models fall to F1 scores of 0.553 on SLC and 0.552 on Targeted data, while Open-weights models drop to 0.492 and 0.530, respectively. This decline happens because, beyond basic retrieval, the need for cross-entity logical reasoning adds an intermediate difficulty that widens the performance gap. The difficulty increases further in Counting tasks. Across all data sources, error rates remain high—such as an NRMSE of 0.255 for Closed-source models and 0.457 for Open-weights models in SLC—highlighting a severe lack of precision. Finally, Calculation tasks represent the highest difficulty, where models, especially Open-weights ones (with RNRMSE approaching 0.491), perform poorly overall, underscoring persistent deficits in current architectures when handling multi-step arithmetic operations.

\section{Conclusion}

We systematically evaluated the capabilities of LLMs in analyzing unstructured text data by building a dedicated social media-based benchmark. Our results indicate that Closed-source LLMs outperform Open-weights models in average performance. However, performance is highly sensitive to question difficulty: while models demonstrate proficiency in simpler Existence questions, they particularly struggle with tasks requiring complex comparison and numerical reasoning. Furthermore, performance also declines as either the semantics or the input schema complexity of the data source increases. 
We further note a universal limitation when input sizes surpass 500 instances: while standard Open-weights models severely deteriorate across tasks, advanced Closed-source models begin to struggle with the numerical aggregation tasks. Consequently, future work should prioritize developing specialized architectures tailored for long-context synthesis and advancing methods to ensure high precision and trustworthy reasoning.

\section*{Limitations}

While our work offers a comprehensive analysis of LLMs in social media data analysis tasks, we acknowledge several inherent limitations shaped by our experimental design and resource constraints.

\paragraph{Data Complexity and Scalability.}
First, the complexity and volume of our dataset were constrained by the high cost of expert annotation. To ensure the reliability of the ground truth, we restricted our scope to data sources derived from classification tasks within the social network domain. Consequently, to maintain manageable human verification standards, the dataset size remains relatively modest. As a result, we have not addressed challenges associated with ``hyper-scale'' analytics, such as the real-time processing of millions of posts, where retrieval latency and context window limitations become critical bottlenecks. Moreover, there may be a risk of contamination by re-purposing existing datasets. Future research should aim to address these data and scalability issues by extending this framework to larger-scale data streams.

\paragraph{Model Scale and Methodological Scope.}
Due to computational budgets and API costs, our evaluation concentrated on cost-effective Closed-source models (e.g., smaller versions), standard Open-weights models, and selected MoE architectures. We were unable to extensively evaluate top-tier proprietary models (e.g., larger ``Pro'' or ``Ultra'' variants), which may possess superior reasoning capabilities. Furthermore, this study focuses on assessing the intrinsic reasoning and calculation capabilities of LLMs via prompting strategies. We did not incorporate agentic workflows utilizing external tools, such as code interpreters or programming agents, despite their growing prevalence in data analysis. While excluding these tools allows for a strict evaluation of the models' native semantic and numerical understanding, it implies that our results likely represent a performance lower bound. Future work should investigate how agentic frameworks and larger-scale proprietary models can further augment performance in this domain. Finally, the fact that we only use English-language dataset prevents us from drawing general conclusions, particularly for low-resource languages.

\section*{Ethical Considerations}
We adhere to the ACL Code of Ethics and ensure that no private or non-public data is used in this work. Our research relies on the widely-used, publicly available datasets TweetEval~\cite{barbieri2020tweeteval} and SuperTweetEval~\cite{antypas2023supertweeteval} and focuses on LLMs for unstructured data analysis. While we use user-generated data from social media (former Twitter), we only report aggregated results over anonymized posts where explicit user mentions were also removed.

For tasks such as spelling and grammar correction, we used privacy-protected AI assistants to ensure confidentiality.


\section*{Acknowledgments}

Yuefeng Shi is supported by the China Scholarship Council (CSC) Grant. Jose Camacho-Collados is supported by a UKRI Future Leaders Fellowship.

\bibliography{custom}

\appendix
\section{Question Data Statistics}
\label{tab:fulldatastas}
Table \ref{tab:dataset_statistics} presents a statistical overview of 
the evaluation benchmarks employed in this study. It details the 
availability of data sample sizes, which range from 10 to 1,000 
instances, and specifies the number of target entities for 
aspect-oriented tasks. Furthermore, the table breaks down the 
distribution of the 470 total questions across three answer formats 
(Yes/No, Number, Percentage) and four logical reasoning categories 
(Calculation, Count, Existence, Comparison).

\begin{table*}[t]
\centering
\small 
\setlength{\tabcolsep}{6pt} 
\begin{tabular}{l c ccc cccc}
\toprule
\multirow{2}{*}{\textbf{Dataset}} & 
\multirow{2}{*}{\textbf{\shortstack{Total\\Qs}}} & 
\multicolumn{3}{c}{\textbf{Answer Type}} & 
\multicolumn{4}{c}{\textbf{Question Type}} \\
\cmidrule(lr){3-5} \cmidrule(lr){6-9} 
 & & \textbf{Y/N} & \textbf{Num} & \textbf{Pct} 
   & \textbf{Calc} & \textbf{Count} & \textbf{Exist} & \textbf{Comp} \\
\midrule
\texttt{TE-Sentiment Analysis}       & 50  & 30 & 10 & 10 & 12 &  8 &  7 & 23 \\
\texttt{TE-Hate Speech Detection}     & 50  & 32 &  9 &  9 & 13 &  5 & 10 & 22 \\
\texttt{TE-Offensive Language Detection}  & 50  & 30 & 10 & 10 & 14 &  6 &  9 & 21 \\
\texttt{STE-Emotion Recognition}                  & 65  & 39 & 14 & 12 & 15 & 11 & 10 & 29 \\
\texttt{STE-Topic Classification}               & 65  & 38 & 13 & 14 & 17 & 10 & 10 & 28 \\
\texttt{STE-Hate Speech Detection}                & 50  & 32 &  9 &  9 & 12 &  6 &  9 & 23 \\
\texttt{STE-Sentiment Analysis}      & 70  & 44 & 13 & 13 & 18 &  8 & 14 & 30 \\
\texttt{TE-Stance Detection}         & 70  & 44 & 13 & 13 & 19 &  7 & 14 & 30 \\
\midrule
\textbf{Total} & 470 & 289 & 91 & 90 & 120 & 61 & 83 & 206 \\
\bottomrule
\end{tabular}
\caption{Detailed statistics of the evaluation datasets.}
\label{tab:dataset_statistics}
\end{table*}

\section{Question Generation Prompts}
\label{sec:Datagenprompts}

\begin{tcolorbox}[
    breakable,
    title={\bfseries Prompt: Evaluation Question Generation},
    colback=white,
    colframe=black,
    arc=3mm,
    boxrule=1.2pt,
    left=2mm,
    right=2mm,
    top=1mm,
    bottom=1mm
]
You are an expert on data analysis for Natural Language Processing (NLP). Your task is to generate a diverse set of candidate evaluation questions based on specific dataset characteristics. The target task configuration is defined as follows:

\vspace{1ex}
\textbf{Task Context:} \texttt{\{Name of the dataset\}} \\
\textbf{Label Set:} \texttt{\{Their corresponding label names\}} \\
\textbf{Label Rules:} \texttt{\{One input has one label or multiple labels\}}

\vspace{1ex}\hrule\vspace{1ex}

\textbf{Generation Directives}

\textbf{1. Core Principles: Real-world Applicability and Grounding}\\
All generated questions must reflect realistic, pragmatically valuable data science inquiries applicable to real-world text analysis scenarios. Strictly avoid overly complex formulations or ``analysis for the sake of analysis'' that lack practical utility. Furthermore, every question must be intrinsically linked to the provided Label Set information, ensuring the analysis is grounded in the dataset's specific content rather than generic metadata.

\textbf{2. Output Quota and Type Constraints}\\
Strictly adhere to a total generation quota of 20 questions, distributed across the following three specific categories:
\begin{itemize}[nosep,left=1em]
    \item \textbf{Binary Questions:} Generate exactly 10 questions that require Yes/No responses.
    \item \textbf{Number Questions:} Generate a subset of numerical questions that require Count/Integer responses. These must be strictly restricted to the positive domain.
    \item \textbf{Percentage Questions:} Generate a subset of numerical questions that require Percentage responses. These must be bounded within the closed interval [0\%, 100\%]. (Note: The combined total of Integer and Percentage questions must equal exactly 10).
\end{itemize}

\textbf{3. Logical Taxonomy and Diversity}\\
The generation process must be conditioned to ensure full coverage of four specific logical categories: \textit{Existence} (Boolean queries regarding the presence of specific label data), \textit{Count} (statistical operations on label instances), \textit{Comparison} (relational analysis between quantities or proportions), and \textit{Calculation} (arithmetic operations, extremes, or proportion analysis). To ensure variety, generate semantic variations using diverse syntactic structures, employing both interrogative and imperative formulations.

\textbf{4. Prohibitions}\\
Do not use hypothetical constructions or conditional terms (e.g., if, given, suppose, assume). All questions must reference the entire dataset directly and must possess exactly one unambiguous answer.

\vspace{1ex}\hrule\vspace{1ex}

\textbf{Output Format}\\
Return the result strictly as a Python dictionary with the following structure:

\begin{ttfamily}
Outputs:\\
\{\\
\hspace*{1em}"Task name": \{\\
\hspace*{2em}"binary": ["generated\_question\_1.", "generated\_question\_2.", ...],\\
\hspace*{2em}"number": ["generated\_question\_1.", "generated\_question\_2.", ...],\\
\hspace*{2em}"percentage": ["generated\_question\_1.", "generated\_question\_2.", ..]\\
\hspace*{1em}\}\\
\}
\end{ttfamily}

\end{tcolorbox}

\section{Question Examples}
\label{sec:fullquestion}
The full question generation examples for each dataset are listed in Table~\ref{tab:comprehensive_examples}.

\begin{table*}[t]
\centering
\scriptsize 
\setlength{\tabcolsep}{4pt} 

\begin{tabularx}{\textwidth}{l l X c}
\toprule
\textbf{Dataset} & \textbf{Type (Ans / Q)} & \textbf{Question} & \textbf{Gold} \\
\midrule

\multirow{6}{*}{\shortstack[l]{\texttt{STE-}\\\texttt{Emotion Recognition}}} 
 & Num / Calc & What is the total count of entries that contain either joy or sadness? & 64 \\
 & Num / Comp & Return the count of instances with the most frequent emotion. & 198 \\
 & Num / Count & How many entries express only one single emotion? & 134 \\
 & Pct / Calc & Calculate the percentage of entries in the dataset that express at least two emotions. & 88.4\% \\
 & Y/N / Comp & Is the combined count of trust and love entries greater than the count of joy entries? & no \\
 & Y/N / Exist & Does every single entry express at least two emotions? & no \\
\cmidrule{1-4}

\multirow{5}{*}{\shortstack[l]{\texttt{STE-}\\\texttt{Hate Speech Detection}}} 
 & Num / Calc & Provide the total number of instances that contain the most prevalent hate type. & 1 \\
 & Num / Count & What is the count of non-hate entries? & 66 \\
 & Pct / Calc & Return the percentage of instances containing hate related to origin. & 4\% \\
 & Y/N / Comp & Does any single hate category make up more than 25\% of all the hateful content? & no \\
 & Y/N / Exist & Does any entry contain multiple types of hate speech? & no \\
\cmidrule{1-4}

\multirow{4}{*}{\shortstack[l]{\texttt{STE-}\\\texttt{Topic Classification}}} 
 & Num / Count & How many entries have the topic `celebrity and pop culture' or `news and social concern'? & 231 \\
 & Pct / Calc & Return the percentage of entries discussing `gaming'. & 4.8\% \\
 & Y/N / Comp & Are entries about the `business' topic the most numerous in the given dataset? & no \\
 & Y/N / Exist & Are there any instances that include both `celebrity and pop culture' and `music'? & yes \\
\cmidrule{1-4}

\multirow{5}{*}{\shortstack[l]{\texttt{TE-Hate}\\\texttt{Speech Detection}}} 
 & Num / Calc & What is the combined count of hateful and non-hateful entries? & 500 \\
 & Num / Count & How many entries express hateful content? & 105 \\
 & Pct / Calc & What is the proportion of hateful entries? & 42\% \\
 & Y/N / Comp & Is the proportion of hateful entries above 30\%? & yes \\
 & Y/N / Exist & Is every entry in the dataset hateful? & no \\
\cmidrule{1-4}

\multirow{5}{*}{\shortstack[l]{\texttt{TE-Offensive}\\\texttt{Language Detection}}} 
 & Num / Calc & What is the absolute difference between offensive and non-offensive entry counts? & 168 \\
 & Num / Count & What is the total number of non-offensive entries? & 502 \\
 & Pct / Calc & Return the percentage number of offensive entries over total dataset. & 33.1\% \\
 & Y/N / Calc & Is the distribution of offensive vs non-offensive entries relatively balanced (within $\pm$10\%)? & no \\
 & Y/N / Exist & Does the dataset include offensive entries? & yes \\
\cmidrule{1-4}

\multirow{5}{*}{\shortstack[l]{\texttt{TE-Sentiment}\\\texttt{Analysis}}} 
 & Num / Calc & What is the total number of positive entries and negative entries? & 55 \\
 & Num / Count & Return the number of entries expressing negative sentiment. & 16 \\
 & Pct / Calc & What percentage does positive data exceed negative data? & 23.5\% \\
 & Y/N / Comp & Are there more positive instances than negative ones in this dataset? & yes \\
 & Y/N / Exist & Does the dataset contain negative sentiment entries? & yes \\
\cmidrule{1-4}
\multirow{5}{*}{\shortstack[l]{\texttt{STE-Sentiment}\\\texttt{Analysis}}} 
 & Num / Calc & What is the total number of positive entries and negative entries towards the target 'kane'? & 24 \\
 & Num / Count & How many entries do not express positive sentiment towards the target ‘ira’ over all entries in the dataset? & 49 \\
 & Pct / Calc & What is the percentage of neutral entries towards the target 'kane'? & 65\% \\
 & Y/N / Comp & Is the ratio of instances with positive sentiment to those as negative sentiment towards the target ‘sharknado’ greater than 2:1? & yes \\
 & Y/N / Exist & Does the dataset contain entries with neutral sentiment towards the target 'sharknado'? & yes \\
\cmidrule{1-4}
\multirow{5}{*}{\shortstack[l]{\texttt{TE-Stance}\\\texttt{Detection}}} 
 & Num / Calc & Return the total number of entries that have a 'favor' stance towards the targets ‘feminist’ or ‘climate’. & 15 \\
 & Num / Count & What is the number of instances that have the 'none' stance towards the target ‘climate’? & 5 \\
 & Pct / Calc &What is the percentage of non-against entries towards the target 'abortion'? & 46\% \\
 & Y/N / Comp & Is the percentage of entries with 'none' or 'favor' stance towards the ‘feminist’ target greater than 25\% over the whole dataset? & no \\
 & Y/N / Exist & Does the dataset contain any entries that do not have the 'none' stance towards the target ‘abortion’? & yes \\

\bottomrule
\end{tabularx}

\caption{Comprehensive evaluation examples across different datasets. The questions cover diverse logical types and answer formats.}
\label{tab:comprehensive_examples}
\end{table*}


\section{Experimental Setup and Implementation Details}

\label{sec:appendix_implementation}

In this section, we provide detailed information regarding the hyperparameters and computational resources used in our evaluation pipeline.

\subsection{Hyperparameters}

To ensure reproducibility and consistency across different model architectures (both Open-weights and Closed-source), we utilized a unified set of generation parameters. Table~\ref{tab:hyperparameters} details the specific configurations used during the inference process.

We employed a low temperature ($T=0.01$) to minimize hallucination and enforce deterministic outputs for data analysis tasks. For local models, we utilized 4-bit quantization via \texttt{bitsandbytes} (NF4 format) to reduce memory overhead while maintaining performance.

\begin{table}[h]
\centering
\small
\begin{tabular}{lr}
\toprule
\textbf{Parameter} & \textbf{Value} \\
\midrule
Temperature & 0.01 \\
Top-p (Nucleus Sampling) & 0.9 \\
Max Response Tokens (Standard) & 20 \\
Quantization (Local Models) & 4-bit (NF4) \\
Compute Dtype & \texttt{float16} \\
Reasoning  Abilities& \texttt{none} \\
\bottomrule
\end{tabular}
\caption{Hyperparameters used for model generation.}
\label{tab:hyperparameters}
\end{table}

\subsection{Experiments Among Different Temperatures}

To determine the temperature for the experiment, we evaluated three representative models, namely GPT-5.4-mini, LLaMA-3.3-70B-Instruct, and Qwen-3.5-397B-A17B ,across four temperature settings (0.01, 0.30, 0.70, and 1.00), aggregating the results over the TE-SA, STE-SA, and STE-ER data source at input sizes of 10, 100, and 1,000 instances. As shown in Table~\ref{tab:temperature}, performance across all three metrics remains highly stable throughout the entire temperature range for every model. Because of no significant performance fluctuations, we elected to use a near-deterministic temperature of 0.01 for our primary experiments. This extremely low setting was chosen specifically to minimize generation variance and ensure that the models strictly adhere to the required output formatting.

\begin{table*}[t]
\centering
\setlength{\tabcolsep}{4pt} 
\resizebox{\textwidth}{!}{
\begin{tabular}{l ccc ccc ccc}
\toprule
 & \multicolumn{3}{c}{\textbf{GPT-5.4-mini}}
 & \multicolumn{3}{c}{\textbf{LLaMA-3.3-70B-Instruct}}
 & \multicolumn{3}{c}{\textbf{Qwen-3.5-397B-A17B}} \\
\cmidrule(lr){2-4}\cmidrule(lr){5-7}\cmidrule(lr){8-10}
\textbf{Temperature}
 & F1$\uparrow$ & RNRMSE$\downarrow$ & NRMSE$\downarrow$
 & F1$\uparrow$ & RNRMSE$\downarrow$ & NRMSE$\downarrow$
 & F1$\uparrow$ & RNRMSE$\downarrow$ & NRMSE$\downarrow$ \\
\midrule
0.01 & 0.734 & 0.339 & 0.252 & 0.733 & 0.505 & 0.514 & 0.764 & 0.302 & 0.277 \\
0.30 & 0.723 & 0.346 & 0.247 & 0.730 & 0.500 & 0.516 & 0.750 & 0.310 & 0.284 \\
0.70 & 0.729 & 0.348 & 0.251 & 0.737 & 0.474 & 0.513 & 0.734 & 0.315 & 0.282 \\
1.00 & 0.735 & 0.352 & 0.253 & 0.729 & 0.438 & 0.512 & 0.735 & 0.306 & 0.287 \\
\bottomrule
\end{tabular}
} 

\caption{Sampling temperature on model performance, aggregated over the TE-SA, STE-SA, and STE-ER datasets (10, 100, 1{,}000 samples each).}
\label{tab:temperature} 
\end{table*}

\section{Task Prompts}
\label{sec:taskprompt}
To ensure rigorous and standardized evaluation across diverse experimental settings, we employ a unified prompt template incorporating dynamic components tailored to specific task requirements. The template integrates a \texttt{\{task\_description\}} (see Table~\ref{tab:task_descriptions})  to characterize the data content and define the analytical domain, thereby activating the model's relevant prior knowledge. 
Furthermore, to guarantee machine readability and structural consistency, the \texttt{\{instructions\}} (see Table~\ref{tab:instructions}) section imposes strict formatting constraints tailored to the specific question type (e.g., enforcing binary "yes/no" labels or raw integer values), thereby minimizing generative noise and ensuring strict adherence to the evaluation protocols.

\begin{tcolorbox}[
    breakable,
    title={\bfseries Prompt Template for Experiments},
    colback=white,
    colframe=black,
    arc=3mm,
    boxrule=1.2pt,
    left=2mm,
    right=2mm,
    top=1mm,
    bottom=1mm
]
You are an expert Text Data Analyst. Your task is to analyze the provided raw text data and answer the user's question with EXACT adherence to the specified format.

\vspace{1ex}
\texttt{\{task\_description\}} {\small\textit{\% Dynamically applied for Instruction Prompts.}} \\

\vspace{1ex}
The data consists of multiple text records, where each record is on a new line. {\small\textit{\% Dynamically applied for Standard inputs.}} \\
The data consists of multiple text records. Each record is an \texttt{<entry>} containing a \texttt{<sentence>} and its corresponding \texttt{<target>} word.{\small\textit{\% Dynamically applied for inputs with target.}}

\vspace{1ex}
--- TEXT DATA BEGIN ---\\
\texttt{\{data\}}\\
--- TEXT DATA END ---

\vspace{1ex}
QUESTION: \texttt{\{question\}}

\vspace{1ex}
CRITICAL FORMAT REQUIREMENTS:\\
\texttt{\{instructions\}}

\vspace{1ex}
WARNING: Any deviation from the specified format will be considered incorrect. Do not provide explanations, reasoning, or additional context.

\vspace{1ex}
ANSWER:
\vspace{1ex}
\end{tcolorbox}

\begin{table*}[t]
\centering
\scriptsize 
\begin{tabularx}{\textwidth}{>{\raggedright\arraybackslash}p{3cm} X}
\toprule
\textbf{Dataset} & \textbf{Task Description Content} \\
\midrule
\texttt{TE-SENTIMENT ANALYSIS} & This dataset has been collected for analysing sentiment, classifying it into three categories: positive, negative, and neutral. \\
\cmidrule(l){1-2}
\texttt{TE-HATE SPEECH} & This dataset has been collected for detecting hate speech, classifying it into two categories, namely hate speech and non-hate speech. \\
\cmidrule(l){1-2}
\texttt{TE-OFFENSIVE LANGUAGE} & This dataset has been collected for analysing offensive language, classifying it into two categories: offensive language and non-offensive language. \\
\cmidrule(l){1-2}
\texttt{STE-EMOTION RECOGNITION} & This dataset has been collected for fine-grained emotion detection analysis, encompassing emotion types such as anger, anticipation, disgust, fear, joy, love, optimism, pessimism, sadness, surprise, trust, among others. \\
\cmidrule(l){1-2}
\texttt{STE-TOPIC CLASSIFICATION} & This dataset has been collected for topic classification analysis, encompassing topic categories such as arts and culture, business and entrepreneurs, celebrity and pop culture, diaries and daily life, family, fashion and style, Film tv and video, fitness and health, food and dining, gaming, learning and educational, music, news and social concern, other hobbies, relationships, science and technology, sports, travel and adventure, youth and student life. \\
\cmidrule(l){1-2}
\texttt{STE-HATE SPEECH} & This dataset has been collected for fine-grained hate speech detection analysis, encompassing hate speech types such as gender, race, sexuality, religion, origin, disability, age, and not hate. \\
\cmidrule(l){1-2}
\texttt{STE-SENTIMENT ANALYSIS} & This dataset has been collected for aspect-based sentiment analysis on a three-point scale, where each entry contains one target for detection and three categories: 'negative', 'neutral', and 'positive'. \\
\cmidrule(l){1-2}
\texttt{TE-STANCE DETECTION} & This dataset has been collected for stance detection, where each entry contains one target for detection and three categories: 'favor', 'against', and 'none'. \\
\bottomrule
\end{tabularx}
\caption{Full list of Task Descriptions. These descriptions are injected into the prompt to provide domain-specific context.}
\label{tab:task_descriptions}
\end{table*}

\begin{table*}[t]
\centering
\small
\begin{tabularx}{\textwidth}{l >{\raggedright\arraybackslash}X >{\raggedright\arraybackslash}X}
\toprule
\textbf{Answer Type} & \textbf{Standard Mode} (\texttt{-{}-with\_explanation=False}) & \textbf{Explanation Mode} (\texttt{-{}-with\_explanation=True}) \\
\midrule
\texttt{yes/no} & 
OUTPUT REQUIREMENT: Respond with exactly ``yes'' or ``no'' only. & 
OUTPUT REQUIREMENT: Respond with ``yes'' or ``no'' on the first line. On a new line, provide a brief explanation for your answer. \\
\cmidrule(l){1-3}

\texttt{number} & 
OUTPUT REQUIREMENT: Respond with only a single integer number (e.g., ``42''). & 
OUTPUT REQUIREMENT: Respond with only a single integer number (e.g., ``42'') on the first line. On a new line, provide a brief explanation. \\
\cmidrule(l){1-3}

\texttt{percentage} & 
OUTPUT REQUIREMENT: Respond with only a percentage in format ``X\%'' (e.g., ``85\%''). & 
OUTPUT REQUIREMENT: Respond with only a percentage in format ``X\%'' (e.g., ``85\%'') on the first line. On a new line, provide a brief explanation. \\
\cmidrule(l){1-3}

\texttt{default} & 
OUTPUT REQUIREMENT: Provide only the direct answer without explanations. & 
OUTPUT REQUIREMENT: Provide only the direct answer on the first line. On a new line, provide a brief explanation. \\
\bottomrule
\end{tabularx}
\caption{Dynamic Format Instructions injected into the \texttt{\{instructions\}} placeholder, contingent on the Answer Type and the presence of the explanation flag.}
\label{tab:instructions}
\end{table*}

\section{Generation Error Statistics}
\label{sec:Generrstas}
Table~\ref{tab:format_validity} reports format validity rates for each dataset and answer type, averaged across all models. Yes/no validity is consistently high across all datasets (93.7–96.7\%), indicating that binary output formatting is rarely a problem regardless of dataset. Number and percentage validity show more variation. SLC datasets have the lowest rates overall, with STE-HS recording the lowest number validity (84.7\%) and percentage validity (77.3\%) in the table. MLC and Targeted datasets are notably higher, with number validity ranging from 91.4\% to 93.7\% and percentage validity from 84.5\% to 86.8\%. Under strict evaluation mode, each invalid output is assigned the worst-case score, so these differences directly affect reported metrics. The strict-mode NRMSE penalty is +0.074 for SLC compared to +0.057 for MLC, meaning part of the cross-category performance gap reflects output formatting rather than prediction quality alone.

\begin{table}[ht]
\centering
\small
\setlength{\tabcolsep}{4pt} 
\begin{tabular}{@{} l c r r r @{}} 
\toprule
\multirow{2}{*}{Dataset} & \multirow{2}{*}{Source} & \multicolumn{3}{c}{Validity Rate (\%)} \\
\cmidrule(l){3-5} 
 & & Y/N & Num. & Pct. \\
\midrule
STE-HS & SLC & 93.7 & 84.7 & 77.3 \\
TE-HS  & SLC & 96.2 & 86.9 & 82.4 \\
TE-OL  & SLC & 95.4 & 88.5 & 83.6 \\
TE-SA  & SLC & 95.9 & 87.9 & 84.9 \\
\addlinespace[2pt]
STE-ER & MLC & 96.3 & 93.7 & 86.1 \\
STE-TC & MLC & 95.5 & 91.4 & 84.5 \\
\addlinespace[2pt]
STE-SA & TC  & 95.6 & 93.1 & 86.8 \\
TE-SD  & TC  & 96.7 & 91.7 & 85.5 \\
\bottomrule
\end{tabular}
\caption{Format validity rates (\% of outputs passing both parsing and range checks) by dataset and answer type (Yes/No, Number, Percentage), averaged across all models and data sizes.}
\label{tab:format_validity}
\end{table}

\section{Detailed Generation Analysis}
\label{app:generation_analysis}

In this section, we provide a granular analysis of the reasoning errors discussed in the main text. Table~\ref{tab:explanation_examples} presents specific output comparisons across three representative models.

\paragraph{Case Study 1: Structural Hallucination (Grok-4.1-Fast).}
As shown in Row 2 of Table~\ref{tab:explanation_examples}, Grok-4.1-Fast constructs a logically sound derivation chain: ``$(23/50) \times 100 = 46\%$''. However, compared to the Gold Answer (84\%), the model hallucinates the total number of entries (claiming 50 instead of the actual count). This demonstrates that the model's alignment training for mathematical reasoning allows it to generate convincing \textit{formats} even when the underlying \textit{premises} are hallucinated.

\paragraph{Case Study 2: Fragmented Reasoning (Qwen-3.5-397B-A17B).}
Qwen-3.5-397B-A17B adopts an evidence-based approach, often explicitly quoting hashtags (e.g., Row 2). However, it exhibits inconsistency in global context tracking. For the same input document where Grok identifies 50 entries (albeit incorrectly), Qwen identifies only 30. This suggests that the MoE architecture's routing mechanism may focus intensely on local text segments at the expense of maintaining a coherent global state, leading to counting discrepancies.

\paragraph{Case Study 3: Refusal and Vague Description (LLaMA-3.3-70B).}
LLaMA-3.3-70B consistently avoids executing complex counting tasks. In Row 2, instead of performing the division, it describes the process (``involves analyzing each text record''). In Row 3, it defaults to a count of ``0'', a safe baseline response when confidence is low. This indicates an alignment strategy prioritizing safety over precision in complex analytical contexts.

\begin{table*}[ht]
\centering
\small

\begin{tabularx}{\linewidth}{@{}p{0.18\linewidth} X X X@{}}
\toprule
\textbf{Question \& Gold Answer} & \textbf{Grok-4.1-Fast} & \textbf{Qwen-3.5-397B-A17B } & \textbf{LLaMA-3.3-70B} \\
\midrule
\textbf{Question:} Are there more entries expressing disgust than any other single emotion? \newline
\textbf{Gold Answer:} no & 
\textbf{Answer:} no \newline
\textbf{Explanation:} Disgust is expressed in at most 2 entries (e.g., ``dreadful'' kit and ``HATE'' ads for meds, implying revulsion), while joy/happiness appears in approximately 18 entries and anger in 7, making disgust less prevalent than several other emotions. & 
\textbf{Answer:} no \newline
\textbf{Explanation:} After analyzing the text data, no entries explicitly express the emotion of disgust. While emotions like fear, sadness, anger, joy, and optimism are clearly present through hashtags... or contextual language, there are no instances of \#disgust or comparable linguistic markers indicating disgust. Therefore, disgust does not occur more than any other single emotion. & 
\textbf{Answer:} no \newline
\textbf{Explanation:} The text data provided contains various emotions such as anger, anticipation, disgust, fear, joy, love, optimism, pessimism, sadness, surprise, and trust. Upon analyzing the data, it appears that emotions like sadness, joy, and anger are more frequently expressed than disgust... Therefore, there are not more entries expressing disgust than any other single emotion. \\
\midrule
\textbf{Question:} Calculate the percentage of entries in the dataset that express at least two emotions. \newline
\textbf{Gold Answer:} 84\% & 
\textbf{Answer:} 46\% \newline
\textbf{Explanation:} The dataset contains 50 entries. After analyzing each for expressed emotions (inferred from text, hashtags, and context...), 23 entries express at least two distinct emotions (e.g., entry 2: fear and anger; entry 20: joy, fear, and relief). Thus, $(23/50) \times 100 = 46\%$. & 
\textbf{Answer:} 35\% \newline
\textbf{Explanation:} Out of 30 text entries, 10 contain clear indicators of at least two distinct emotions (e.g., \#nervous and [flushed face] in one entry indicating anxiety and surprise; \#depression, \#failure, \#worthless showing sadness, self-loathing, and pessimism). These were identified through explicit emotion tags... $10/30 \approx 33.3\%$, which rounds to 35\% when accounting for borderline cases. & 
\textbf{Answer:} 47\% \newline
\textbf{Explanation:} The calculation of the percentage of entries expressing at least two emotions involves analyzing each text record for emotional cues... The total number of records expressing at least two emotions is then divided by the total number of records to obtain the percentage. This process requires a detailed examination of linguistic patterns... \\
\midrule
\textbf{Question:} Count entries containing at least four different emotions. \newline
\textbf{Gold Answer:} 2 & 
\textbf{Answer:} 2 \newline
\textbf{Explanation:} The text data consists of 10 entries. After fine-grained emotion detection, entries 4 (anger, sadness, trust, optimism) and 10 (sadness, disgust, anger, pessimism) each contain at least four distinct emotions from the specified set... All other entries express fewer than four. & 
\textbf{Answer:} 1 \newline
\textbf{Explanation:} One entry contain at least four different emotions, which is ``\& like if your unhappy just leave...'' --- it expresses unhappiness, implied anger, betrayal, and reference to love, totaling at least four. & 
\textbf{Answer:} 0 \newline
\textbf{Explanation:} The provided text data does not explicitly contain entries with labeled emotions, but rather implicit expressions of emotions through text. Upon analyzing the text, it appears that most entries express one or two dominant emotions, but none of the entries explicitly contain at least four different emotions. Therefore, the count of such entries is 0. \\
\bottomrule
\end{tabularx}
\caption{Examples of Question Explanation Generation by different models compared to Gold Answers. }
\label{tab:explanation_examples}
\end{table*}

\end{document}